\pdfoutput=1 
\documentclass{article}  
\usepackage{cite}
\usepackage{calc}
\usepackage{graphicx}
          \begin{document} 

		\title{Information and search in computer chess }   
		\author{Alexandru Godescu, ETH Zurich}
		\maketitle

	\newtheorem{axiom}{Axiom}
	\newtheorem{theorem}{Theorem}
	\newtheorem{definition}{Definition}	
	\newtheorem{proposition}{Proposition}
	\newtheorem{interpretation}{Interpretation}
	\newtheorem{analysis}{Analysis}
	\newtheorem{search experiment}{Search Experiment}
	\newtheorem{assumption}{Assumption}
	\newtheorem{principle}{Principle}
	\newtheorem{conjecture}{Conjecture}
	\newtheorem{positional analysis}{positional analysis}
	\newtheorem{positional judgment }{positional judgment}
	\newtheorem{experimental evidence}{experimental evidence}

	\begin{abstract}

		The article describes a model of chess based on information theory. A mathematical model of the partial depth scheme is outlined and
		a formula for the partial depth added for each ply is calculated from the principles of the  model. 
		An implementation of alpha-beta with partial depth is given . 
		The method is tested using an experimental strategy having as objective to show the effect of allocation of a higher
		amount of search resources on areas of the search tree with higher information. The search proceeds in the direction of lines with higher 				information gain. The effects on search performance of allocating higher search resources on lines with higher information gain are 
		tested experimentaly and conclusive results are obtained. In order to isolate the effects of the partial depth scheme no other heuristic is used.

	\end{abstract}

	\newpage

	\section{Introduction}

	\subsection{ Motivation}

There is  gap in the scientific analysis of the fraction ply methods one of the best methods of search in computer chess and other strategy games.
As Hans Berliner pointed out about the scheme of ''partial depths'', ''...the success of these micros (micro-processor based programs) attests to the efficacy of the procedure. Unfortunately, little has been published on this''. This research has the objective of developing a theoretical model of the partial depth scheme based on information theory, implementing it and providing experimental evidence for the method and for the model.
	
	\subsection{The research methodology and  scenario}

	An introduction to games theory and information theory is given in the background section. A model  based on the principles of information theory is outlined and then the formula for partial depths scheme is calculated. Search experiments are performed and then the results are interpreted. In the appendix can be found an introduction to some concepts in chess, and to the axioms of information theory.

	\subsection{Background knowledge}

	\subsubsection{The games theory model of chess}

	An important mathematical branch for modeling chess is games theory, the study of strategic interactions.
	
	\begin{definition}
		Assuming the game is described by a tree, a finite game is a game with a finite number of nodes in its game tree.
	\end{definition}
	
	It has been proven that chess is a finite game. The rule of draw at three repetitions and the 50 moves rule ensures that chess is a finite game. 

	\begin{definition}
		Sequential games are games where players have some knowledge about earlier actions. 
	\end{definition}

	\begin{definition}
		A game is of perfect information if all players know the moves previously made by all players. 
	\end{definition}

	Zermelo proved that in chess either player $\romannumeral 1$ has a winning pure strategy, player $\romannumeral 2$ has a winning pure strategy, or either player can force a draw.

	\begin{definition}
		A zero sum game is a game where what one player looses the other wins.
	\end{definition}

	Chess is a two-player, zero-sum, perfect information game, a classical model of many strategic interactions. 

	By convention, W is the white player in chess because it moves first while B is the black player because it moves second.
	Let M(x) be the set of moves possible after the path x in the game has been undertaken.
	W choses his first move $w_1$ in the set M of moves available.  B chooses his move $b_1$ in the set M($w_1$): $b_1$  $\in$ M($w_1$)
	Then W chooses his second move $w_2$, in the set M($w_1$,$b_1$): $w_2$  $\in$  M($w_1$,$b_1$) 
	Then B chooses his his second move $b_2$ in the set M($ w_1$,$b_1$,$w_2$): $b_2$ $\in$ M($w_1$,$b_1$,$w_2$) At the 			end, W chooses his last move $w_n$ in the set M($w_1$, $b_1$, ... ,$w_{n-1}$ ,$b_{n-1}$  ). \newline
	In consequence $w_n$ $\in$ M($w_1$, $b_1$, ... ,$w_{n-1}$ ,$b_{n-1}$  )

	Let n be a finite integer and M, M($w_1$), M($w_1$,$b_1$),...,\linebreak M($w_1$, $b_1$, ... ,$w_{n-1}$ ,$b_{n-1}$,$w_n$) 
	be any successively defined sets for the moves $w_1$,$b_1$,...,$w_n$,$b_n$ satisfying the relations: \newline
	
	\begin{equation}  b_n  \in  M(w_1, b_1, ... ,w_{n-1} ,b_{n-1},w_n  )  \label{EQ} \end{equation}    and \newline
	\begin{equation}  w_n  \in  M(w_1, b_1, ... ,w_{n-1} ,b_{n-1}  )         \label{EQ} \end{equation}    \newline   

	\begin{definition}
		A realization of the game is any 2n-tuple ($w_1$, $b_1$, ... ,$w_{n-1}$ ,$b_{n-1}$,$w_n$,$b_n$ ) satisfying the relations (1) and (2) 	
	\end{definition}

	A realization is called variation in the game of chess.

	Let R be the set of realizations (variations) , of the chess game. Consider a partition of R in three sets $R_w$ ,$R_b$ and $R_{wb}$ so that for any realization in $R_w$, player1 ( white in chess )  wins the game, for any realization in $R_b$ , player2 (black in chess) wins the game and for any realization in $R_{wb}$, there is no winner (it is a draw in chess).

	Then R can be partitioned in 3 subsets so that
	
	\begin{equation} R = R_w + R_b + R_{wb}   \label{EQ} \end{equation}    \newline  

	W has a winning strategy if $\exists$   $w_1$  $\in$   M ,  $\forall$  $b_1$  $\in$  $M(w_1)$  ,\linebreak $\exists$  $w_2$ $\in$ M($w_1$,$b_1$)  ,   $\forall$  $b_2$ $\in$ M($ w1$, $b1$, $w2$ ) ...\newline $\exists$ $w_n$ $\in$ M($w_1$,$b_1$,...,$w_{n-1}$,$b_{n-1}$), \linebreak $\forall$ $b_n$ $\in$ M($w_1$,$b_1$,...,$w_{n-1}$,$b_{n-1}$,$w_n$) ,
  where the variation
\begin{equation} ( w_1 , b_1 , \ldots , w_n, b_n ) \in R_w  \label{EQ}\end{equation} 

   	W has a non-loosing strategy if $\exists$   $w_1$  $\in$   M ,  $\forall$  $b_1$  $\in$  $M(w_1)$  ,\linebreak $\exists$  $w_2$ $\in$ M($w_1$,$b_1$)  ,   $\forall$  $b_2$ $\in$ M($ w_1$, $b_1$, $w_2$ )...\newline  $\exists$ $w_n$ $\in$ M($b_1$,$w_1$,...,$w_{n-1}$,$b_{n-1}$), \linebreak $\forall$ $b_n$ $\in$ M($w_1$,$b_1$,...,$w_{n-1}$,$b_{n-1}$,$w_n$) , where the variation
\begin{equation} ( w_1 , b_1 , \ldots , w_n, b_n ) \in R_w + R_{wb} \label{EQ}\end{equation}

	B has a winning strategy if $\exists$   $b_1$  $\in$   M ,  $\forall$  $w_1$  $\in$  $M(w_1)$  ,\linebreak $\exists$  $b_2$ $\in$ M($w_1$,$b_1$,$w_2$ )  ,   $\forall$  $w_2$ $\in$ M($ w_1$, $b_1$) ...\newline  $\exists$ $b_n$ $\in$ M($w_1$,$b_1$,...,$w_{n-1}$,$b_{n-1}$,$w_n$), \linebreak $\forall$ $w_n$ $\in$ M($w_1$,$b_1$,...,$w_{n-1}$,$b_{n-1}$) ,
  where the variation
\begin{equation} ( w_1 , b_1 , \ldots , w_n, b_n ) \in R_b  \label{EQ}\end{equation} 

   	B has a non-loosing strategy if $\exists$   $b_1$  $\in$   M ,  $\forall$  $w_1$  $\in$  $M(w_1)$  ,\linebreak $\exists$  $w_2$ $\in$ M($w_1$,$b_1$)  ,   $\forall$  $w_2$ $\in$ M($ w_1$, $b_1$) ...\newline  $\exists$ $b_n$ $\in$ M($w_1$,$b_1$,...,$w_{n-1}$,$b_{n-1}$,$w_n$), \linebreak $\forall$ $w_n$ $\in$ M($w_1$,$b_1$,...,$w_{n-1}$,$b_{n-1}$) , where the variation
\begin{equation} ( w_1 , b_1 , \ldots , w_n, b_n ) \in R_b + R_{wb} \label{EQ}\end{equation}

\begin{theorem}  
	Considering a game obeying the conditions stated above, then each of the next three statements are true:\newline
	(\romannumeral 1). W has a winning strategy or B has a non-losing strategy. \\
	(\romannumeral 2). B has a winning strategy or W has a non-losing strategy. \\
	(\romannumeral 3). If $R_{wb}$ = $\emptyset$, then W has a winning strategy or B has a winning strategy.   \\
\end{theorem}
 
	If $R_{wb}$ is $\emptyset$, one of the players will win and if  $R_{wb}$ is identical with R the outcome of the game will result in a draw at perfect play from both sides. It is not know yet the outcome of the game of chess at perfect play.\\

	The previous theorem proves the existence of winning and non-losing strategies, but gives no method to find these strategies. A method would be to transform the game model into a computational problem and solve it by computational means.  Because the state space of the problem is very big, the players will not have in general, full control over the game and often will not know precisely the outcome of the strategies chosen. The amount of information 
gained in the search over the state space will be the information used to take the decision. The quality of the decision must be a function
of the information gained as it is the case in economics and as it is expected from intuition.

	\subsubsection{Concepts in information theory}
	
	Of critical importance in the model described is the information theory. It is proper to make a short outline of information theory 			concepts used in the information theoretic model of strategy games and in particular chess and computer chess. 
	
	\begin{definition}
		A discrete random variable $\chi$ is completely defined by the finite set of values it can take S, and the probability 			distribution ${P_x(x)}_{x \in S}$. The value  $P_x(x)$ is the probability that the random variable $\chi$ takes the value x.
	\end{definition}
	
	\begin{definition}
The probability distribution $P_x$ :S $\rightarrow$ [0,1] is a non-negative function that satisfies the normalization 			condition \begin{equation}  \sum_{x \in S} P_x(x) = 1   \label{EQ}\end{equation}
	\end{definition}

	\begin{definition}
		The expected value of f(x) may be defined as 
		\begin{equation}  \sum_{x \in S} P_x(x)*f(x)   \label{EQ}\end{equation}
	\end{definition}

	This definition of entropy may be seen as a consequence of the axioms of information theory. It may also be defined independently ~\cite{CoverThomas}. As a place in science and in engineering, entropy has a very important role. Entropy is a fundamental concept of 
the mathematical theory of communication, of the foundations of thermodynamics, of quantum physics and quantum computing. 

	\begin{definition}
		The entropy $H_x$ of a discrete random variable $\chi$ with probability distribution p(x) may be defined as
		\begin{equation} H_x = - \sum_{x \in S} {p(x)*\log{ p(x)}}   \label{EQ}\end{equation}
	\end{definition}

	Entropy is a relatively new concept, yet it is already used as the foundation for many scientific fields. This article creates the 
	foundation for the use of information in computer chess and in computer strategy games in general. However the concept
	of entropy must be fundamental to any search process where decisions are taken.

	Some of the properties of entropy used to measure the information content in many systems are the following:

	\paragraph{ Non-negativity of entropy }

	\begin{proposition} 
		\begin{equation}   H_x \geq 0 \label{Prop}\end{equation}
	\end{proposition}

	\begin{interpretation} 
		Uncertainty is always equal or greater than 0.If the entropy, H is 0, the uncertainty is 0 and the random variable x takes a certain value with probability $P(x)$ = 1   
	\end{interpretation}

	\begin{proposition} 
		Consider all probability distributions on a set S with m elements.
	           H is maximum if all events x have the same probability, $p(x)$ = $\frac{1}{m}$ 	
	\end{proposition}

	\begin{proposition} 
		If X and Y are two independent random variables , then 
		\begin{equation}     P_{X,Y}(x,y) = P_x(x)*P_y(y)           \label{Prop}\end{equation}
	\end{proposition}

	\begin{proposition} 
		The entropy of a pair of variable X and Y is \begin{equation}  H_{x,y} = H_x + H_y  \label{Prop}\end{equation}
	\end{proposition}

	\begin{proposition} 
		For a pair of random variables one has in general 
		\begin{equation} H{x,y} \leq  H_x + H_y \label{Prop}\end{equation}
	\end{proposition}

	\begin{proposition} Additivity of composite events

		The average information associated with the choice of an event x is additive, being the sum of the information
		associated to the choice of subset and the information associated with the choice of the event inside the subset, 							weighted by the probability of the subset		

	\end{proposition}

	\begin{definition}
		
		The entropy rate of a sequence   $x_N$ =   $X_t$ , t $\in$ N
		\begin{equation} h_x = \lim_{N\to\infty}\frac{H_{x_N}}{N}   \label{def}\end{equation}

	\end{definition}

	\begin{definition}

	Mutual information is a way to measure the correlation of two variables

\begin{equation} I_{X,Y} = - \sum_{x \in S, y \in T} {p(x,y)*\log{ \frac{ p(x,y) }{  p(x)*p(y) }  }  }	\label{def}\end{equation}

	\end{definition}

	All the equations and definitions presented have a very important role in the model proposed as will be seen later in the article.

	\begin{proposition}

	\begin{equation}	I_{x,y} \geq 0	\label{def}\end{equation}

	\end{proposition}

	\begin{proposition}

		\begin{equation}	I_{X,Y} = 0 \label{def}\end{equation}   if any only if X and Y are independent variables.

	\end{proposition}

	\subsection{Previous research in the field}

A necessary condition for a truly selective search given by Hans Berliner is the following : The search follows the areas with highest information in the tree ~\cite{HBerliner2}  ``It must be able to focus
	the search on the place where the greatest information can be gained toward terminating the search''.
	Berliner describes the essential role played by information in chess, however he does not formalize the concept of information
	in chess as an information theoretic concept. From the perspective of the depth in understanding the decision process in chess
	the article  ~\cite{HBerliner2} is exceptional but it does not formulate his insight in a mathematical frame. It contains great chess and computer chess analysis but it does not define the method in mathematical definitions, concepts and equations.	

 Mark Winands in ~\cite{MWinands} outlines a method based on fractional depth where the fractional ply FP of a move with a category c is given by 
	
		\begin{equation}    FP = \frac{\lg P_c}{ \lg C }   \label{EQ}\end{equation}

	His approach is experimental and based on data mining as the method presented previously.

 In the article ~\cite{SEX}  David Levy, David Broughton, Mark Taylor describe

	the selective extension algorithm. The method is based on ''assigning an appropriate additive measure for the interestingness of the terminal node'' of a path.

	Consider a path in a search tree consisting of the moves $M_1$, $M_{ij}$, $M_{ijk}$ and the resulting position being a 
	terminal node. The probability that a terminal node in that path is in the principal continuation is

	\begin{equation}   P( M_i )*P( M_{ij} )*P( M_{ijk} )     \label{EQ}\end{equation}

	The measure of the ''interestingness'' of a node in this method is

	 \begin{equation}      lg[  P( M_i ) ] + lg[ P( M_{ij}  ] + lg[ P( M_{ijk} )  ]     \label{EQ}\end{equation}

	\subsection{analysis of the problem}

	The problem is to describe the mathematical meaning of information in computer chess,  develop the principles and formulas that can be used to 
	control the search and provide experimental evidence for the search heuristic as well as for the role of information gain in obtaining good results
	at an acceptable cost.  
	
	\subsection{Contributions}

The contributions of this research are the creation of the information theoretical model for search in computer chess, the description of the information gain in computer chess and a scientific explanation of the partial depth scheme. The paths explored are the areas of the search tree with the highest amount of information gain. Other contributions are, the calculation of information gain for important moves, the calculation of a formula describing the size of the ply added for various moves, the experimental evidence given for the effect of information gain on search for chess problems. 

	\section{Search and decision methods in computer chess}

	\paragraph{Search on informed game trees \newline} 

	In ~\cite{PijlsBruin} it is introduced the use of heuristic information in the sense of upper and lower bound but no reference to any information theoretic concept is given. Actually the information theoretic model would consider a distribution not only an interval as in ~\cite{PijlsBruin}.
	Wim Pijls and Arie de Bruin presented a interpretation of heuristic information based on lower and upper estimates for a node and integrated it in alpha beta, proving in the same time the correctness of the method under the following specifications.  

Consider the specifications of the procedure alpha-beta.
If the input parameters are the following: \newline
(1) n, a node in the game tree,               \newline
(2) alpha and beta , two real numbers and  \newline
(3) f , a real number, the output parameter, \newline
and the conditions: \newline
(1)pre:   alpha $<$ beta  \newline
(2)post: \newline	
	alpha $<$ f $<$ beta   $\Longrightarrow f = f(n)$,                         \newline
	f $\leq$  alpha	$\Longrightarrow$ f(n) $\leq$  f $\leq$  alpha  \newline
	f $\geq$  beta  $\Longrightarrow$ f(n) $\geq$ f $\geq$ beta          \newline

then

\begin{theorem}

	The procedure alpha-beta (defined with heuristic information, but not quantified as in information theory)   meets the specification.~\cite{PijlsBruin}

\end{theorem}

	Considering the representation given by  ~\cite{PijlsBruin}, assume for some game trees, heuristic information on the minimax value f(n) is available
	for any node. 

          \begin{definition}
	              The information may be represented as a pair H = (U,L), where U and L map nodes of the tree into real numbers.
          \end{definition}

	\begin{definition}
	           U is a heuristic function representing the upper bound on the node. 
	\end{definition}

	\begin{definition}
	           L is a heuristic function representing the lower bound on the node.
	\end{definition}

	For every internal node, n the condition U(n) $\geq$ f(n) $\geq$ L(n) must be satisfied. \newline

	For any terminal node n the condition U(n) = f(n) = L(n) must be satisfied. This may even be considered as a condition for a leaf.

	\begin{definition}
		A heuristic pair H = (U,L) is consistent if  \\
	    	U(c) $\leq$ U(n)      for every child c of a given max node n and  \\
	  	L(c) $\geq$ L(n)      for every child c of a given min node n 
	\end{definition}

	The following theorem published and proven by ~\cite{PijlsBruin} relates the information of alpha-beta and the set of nodes visited.

	\begin{theorem}

		Let $H_1$ = ($U_1$,$L_1$) and $H_2$ = ($U_2$,$L_2$) denote heuristic pairs on a tree G, such that $U_1$(n) $\leq$ $U_2$(n) and $L_1$(n) $\geq$ $L_2$(n) for any node n. Let $S_1$ and $S_2$ denote the set of nodes, that are visited during execution of the alpha-beta procedure on G with $H_1$ and $H_2$ respectively, 
then $S_1$ $\subseteq$ $S_2$.

	\end{theorem}

	\subsection{Problem formulation}

	In the light of the new description it is possible to reformulate the search problem in a strategy game.
	The problem is to plan the search process minimizing the entropy on the value of the starting position considering limits in costs. The best case is when entropy, or uncertainty in the 			value of a position becomes 0 with an acceptable cost in search. This is feasible in chess and it happens every time when a successful combination is executed and results in mate or significant advantage. 

	It is possible to formulate the problem of search in computer chess and in other games as a problem of entropy minimization.

	\begin{equation} Min \{ H(position)  \} = Min \{ -\sum_{i=1}^{\infty}{P_i}log P_i \}  \label{EQ}\end{equation}

	subject to a limit in the number of position that can be explored.

	\section{The model}

	\begin{assumption}

The entropy of a position can be approximated by the sum of entropy rates of the pieces minus the entropy reduction due the strategical configurations. 
\end{assumption}

This can be expressed as:

	\begin{equation} H_{trajectory}(position) 	=  \sum_{i=1}^{N} H_{p_i} -  \sum_{i} H_{s_i}   \label{EQ}\end{equation} 

where $H_i$ represents the entropy of a piece and $H_{s_i}$ represents the entropy of a structure with possible strategic importance. 

This gives also a more general perspective on the meaning of a game piece. A game piece can be seen as a stochastic function having the state of the board as entrance and 
generating possible trajectories and the associated probabilities. These probabilities form a distribution having an uncertainty associated. 

The entropy of a positional pattern, strategic or tactic may be considered a form of joint entropy of the set of variables represented by pieces positions and their trajectory.
The pieces forming a strategic or tactic pattern have correlated trajectories which may be considered as forming a plan.

\begin{equation}H(s_i) =  - \sum_{x_i} ... \sum_{x_n} P(s_i) \log [P(s_i)]	 \label{EQ}\end{equation}

\begin{equation} H_{s_i} =  H(s_i) \label{EQ}\end{equation}

where $s_i$ is a subset of pieces involved in a strategic pattern and the probabilities $P_i$ represent the probability of realization of such strategic or tactical pattern. The reduction of entropy caused by strategic and tactical patterns such as double attacks,pins, is determined by both the frequency of such 
structures and by a significant increase in the probability that one of the sides will win after this position is realized.

We may consider the pieces undertaking a common plan as a form of correlated subsystems with mutual information I(piece1,piece2,...). It results
that undertaking a plan may result in a decrease in entropy and a decrease in the need to calculate each variation. It is known from practice that planning decreases the need to calculate each variation and this gives an experimental indication for the practical importance of the concept of entropy as it is defined here in the context of chess . Each of the tactical procedures , pinning, forks, double attack, discovered attack and so on, can be understood formally in this way. A big reduction in the uncertainty in regard to the outcome of the game 
occurs, as the odds are often that such a structure will result in a decisive gain for a player.
When such a structure appears as a choice it is likely that a rational player will chose it with high probability. 

The entropy of these structures may be calculated with a data mining approach to determine how likely they appear in games.

An approximation if we do not consider the strategic structures would be: 

\begin{assumption}                                     
 
	\begin{equation} H_{trajectory}(position) 	=  \sum_{i=1}^{N} H_{p_i}  \label{EQ}\end{equation} 

\end{assumption}

\paragraph{assumption analysis:} The entropy of the position is smaller in general than the sum of the entropies of pieces because there are certain positional patterns such as openings, end-games, various pawn configurations  in a chess position which result in a smaller number of combinations, results in order and a smaller entropy. Closer to reality would be such a statement:

	\begin{equation} H_{trajectory}(position) 	\leq  \sum_{i=1}^{N} H_{p_i}  \label{EQ}\end{equation}

	\section{RESULTS}

	\subsection{A definition of information gain in computer chess}

	It is possible to define the information gain during the search process based on the reduction in uncertainty in the following way:

	\begin{equation}    I_{gain}  =  \triangle H         \label{EQ}\end{equation}

	Where H represents the uncertainty in the value of the position and  $\triangle H$    

	\begin{equation}	\triangle H = H_2 - H_1   \label{EQ}\end{equation}

	represents the variation of uncertainty in the current position after a move is made. It is the information gained after making a move.

	In the case when \begin{equation}  \triangle H  \le 0 \label{EQ}\end{equation}  we speak of information gain, 

	if \begin{equation} \triangle H \ge 0 \label{EQ}\end{equation}  we understand information lost through approximate evaluation or other operation.

It is possible to describe the information gain of the search process by defining the heuristic efficiency

	\begin{equation}	HE =  \frac{I_{gain}}{\triangle Nodes}  = \frac{\triangle H}{\triangle Nodes}      \label{EQ}\end{equation}

	When   $\triangle  Nodes$   $\longrightarrow$ 1 the information gain results after a move is

	\begin{equation}    I_{gain}(Move) =  H(before Move) - H(after Move)         \label{EQ}\end{equation}

	This concept may be considered similar to the the concept of information gain for decision trees, the Kullback-Leibler divergence.
	We may see the same principle also here, the higher the difference between entropies, the higher the information gain, which makes very 			much sense also intuitively and it provides a new theoretical justification for the empirical heuristics of chess and computer chess.

	\subsection{The justification of the partial depths scheme using the information theoretic model}

	The partial depths method is a generalization of the classic alpha beta in that it offers a greater importance to moves considered significant
	for the search. If all moves have the same importance then , the partial depth scheme can be reduced to the ordinary alpha-beta scheme.
           It can be described also as an importance sampling search. The partial depth scheme has been used by various authors. As Hans
	Berliner observed, few has been published about this method   ~\cite{HBerliner2}.The contribution of this article goes in this direction.

	It is possible to define a function returning the depth:

	\begin{equation}	\triangle depth = f( path )   \label{EQ}\end{equation}

	This is a generalization of the classic alpha-beta because in classic alpha-beta $\triangle$ depth $=$ constant;
	If the decision to add a certain depth to the path is dependent only on the current move and position , 
then if $\triangle$ path ${\longrightarrow}$ 1
	the decision depends only on the current position.

	The increase in depth is dependent on the path in this method, where the path is composed of moves  $m_1$ ,$m_2$ , $m_3$ , .... .  In the classic alpha-beta the depth increase is constant regardless of the type of move.

	\subsection{The design of a search method with optimal cost for a certain information gain}

	The principle behind a theory of optimal search should be the allocation of search resources based on the optimality of information gain per cost.
	It results that the fraction of a search ply added to the depth of the path with a move should be in inverse proportion to the quality of the move.
	The standard approach gives equal importance to all moves, the fraction ply method gives more importance to significant moves.
	Therefore it must be described a quantitative measure for the quality of a move. The reduction from the normal depth of 1 ply should be 
	proportional to the quantitative measure of the quality of a move.

	The fraction ply FD must decrease with the quality of the move relative to optimal. The fraction ply added would be equal in this system to the decrease of a full play with the approximate entropy reduction achieved by that move compared
	to a move having the highest entropy reduction.

	For instance for a capture of a rock the entropy reduction is $\log 14$ 

	\begin{axiom}
	
	An axiom of efficient search in chess , in computer chess and of efficient search in general should be that the probability of executing a move must be equal to the heuristic efficiency of that move which is equal to the information efficiency of expanding the node resulted after the move. The same principle can be considered in general for trajectories.

	\end{axiom}

	By notation, let the heuristic efficiency be HE and   $P_{c_i}$ be the probability of a move in category   $c_i$ to be executed. 
	The heuristic efficiency is a fundamental measure of the ability of a search procedure to gain information from the state space. The heuristic efficiency depends in this
	analysis on the categories of moves and trajectories defined. The examples are for moves with individual tactical values, however the analysis can be extended also
	to tactical plans generated by pins, forks and other tactical patterns. Because such analysis would require some readers to look for the meaning of these structures in 			chess books and also because space considerations the moves generating such configurations would not be presented as examples.  No additional theoretical      	difficulties would emerge from the introduction of these move categories. The same applies to strategical elements. 

	Following the principles outlined, a formula for the fraction ply can be derived.

	\begin{equation}	P_{c_i} = k* HE  \label{EQ}\end{equation}

	Considering that 

	\begin{equation}   HE = \frac{  I_{gain_i} }  { \triangle Cost  }  \label{EQ}\end{equation}

	and

	\begin{equation} \frac{ I_{gain_i} }{ \triangle Cost } = \frac {\triangle Entropy_{category_i } } { \triangle Cost  }    \label{EQ}\end{equation}

	it means 

	\begin{equation} HE = \frac{\triangle Entropy_{category_i} } { \triangle Cost  } \label{EQ}\end{equation}

	For k = 1, \begin{equation}  P_{c_i} =  \frac{\triangle Entropy_{category_i} } { \triangle Cost  } \label{EQ}\end{equation}

	from this, \begin{equation}   \triangle Cost = \frac{\triangle Entropy_{category_i} } { P_{c_i}  }  \label{EQ}\end{equation}

	Of course a different value than 1 can be given to the constant k and this will propagate without changing the meaning of the equations. The constant k would increase the flexibility of implementations actually, offering more freedom in this direction. 
	Now consider the same equation for the move category with the best information gain.

	It means \begin{equation}  P_{c_{BestGain} } =  \frac{\triangle Entropy_{BestGain} } { \triangle Cost  } \label{EQ}\end{equation}

	Assuming the moves from the best category, the most informational efficient will always be executed in the search, the following condition must be satisfied: 

          \begin{equation}  P_{c_{BestGain} } = 1 \label{EQ}\end{equation}

	Then  	\begin{equation}	\frac{\triangle Entropy_{BestGain} } { \triangle Cost  } = 1   \label{EQ}\end{equation}

	so

	\begin{equation}  \triangle Cost = \triangle Entropy_{BestGain} \label{EQ}\end{equation}
	
	The cost for execution of any of the two moves is the same. Equating this cost, it results

	\begin{equation} \frac{\triangle Entropy_{category_i} } { P_{c_i}  } = \triangle Entropy_{BestGain}  \label{EQ}\end{equation}

	It means 

	\begin{equation}  P_{c_i} = 	\frac{\triangle Entropy_{category_i} } { \triangle Entropy_{BestGain} } \label{EQ}\end{equation}

	which is a very intuitive result.

	In general, for a $trajectory_i$, the probability of a trajectory to be explored should be in this system
	
	\begin{equation}  P_{trajectory_i} = 	\frac{\triangle Entropy_{trajectory_i} } { \triangle Entropy_{BestTrajectory} } * \frac{\triangle Cost_{BestTrajectory}}{\triangle Cost_{trajectory_i}} \label{EQ}\end{equation}

	\subsection{The ERS* , the entropy reduction search in computer chess}

	Let $ P_{c_i}$  be the probability that a move is executed and one more ply is added to the search. 

	The size of the ply added should be function of this probability. It is logically to consider the size of the play as a quantity increasing with the
	probability of the move not being executed. The probability of the move not being executed is $1 - P_{c_i}$ therefore
	assuming an abstraction, a linear relation of the form:\newline size of ply = k*( probability of  a move not being executed )
	then the relation between the size of the ply and the probability of the move to be chosen would be for k = 1

	         \begin{equation}   D = 1 - \frac{\triangle Entropy_{category_i} } { \triangle Entropy_{BestGain} }   \label{EQ}\end{equation}

	This may be considered even a theorem describing the size of the fraction ply in computer chess and even for other EXPTIME problems under the above assumptions and resulting from the above calculations.

	Starting from the previous equation, it is possible to use the relative entropies of pieces and positional patterns to implement the previous 			formula.

	 Consider the check as the move with the ultimate decrease in entropy because its forceful nature and because it has a higher frequency in the vicinity of the objective, the mate than any other move.  Then all the other moves may be rated as function of the check move. Let such value be $\log 30$ . Here can be used a constant 
reflecting the above mentioned properties of such move. It must be noted that not all checks are equally significant. Several categories of checks can be introduced instead of 
a single check category. Also in the application, not all checks are equally important, check and capture for example gains a better priority but in this example does not have a smaller depth. 
		
	As a consequence, if the normal increase in search depth is counted as 1 for moves without significance the fractional ply for a check is:

	\begin{equation}  D = 1  - \frac {\log 30} {\log 30}	\label{EQ}\end{equation}

	then D = 0 in this system because the best move should be always executed and then the depth added should be 0.

	For a capture of queen the entropy rate of the system decreases with $\log 28$.

	Then the fractional ply for a queen capture is 

	\begin{equation}  D = 1  - \frac {\log 28} {\log 30}	\label{EQ}\end{equation}
	
	after calculations, D = 0.02

	For a capture of rock the entropy rate of the system decreases with $\log 14$.

	Then the fractional ply for a rock capture is

	\begin{equation}  D = 1  - \frac {\log 14} {\log 30}	\label{EQ}\end{equation}
	
	after calculations, D = 1 - 0.776 = 0.223

	Instead of using the entropy rates for calculating the size of the fractional depth it is possible to use the value of pieces which is strongly correlated
	for most of the systems with the entropy rate of the pieces.As it can be seen from the calculation above, the higher the differences in entropy
	between consecutive positions in a variation, the higher the information gain. This can be understood as a divergence between distributions
	of consecutive moves. The more they diverge the higher the information gain after a move.

	\subsection{ Experimental results }

	 As a test case it is used a combination which gives us the possibility to define the quality of the response to a position in a precise way.
	The meaning of the columns is the following:\\
	column 1:EXPERIMENT NUMBER  - represents the number of the search experiment \\
	column 2:NODES SEARCHED  - represents the number of nodes searched in the experiment  \\
	column 3:TERM DIVIDING THE REDUCTION IN PLY - represents the number dividing the term decreasing the size of the normal ply added to the current depth \\ 
           column 4: MAX DEPTH ATTAINED - the maximum depth in standard plies attained , here it is added 1 for each ply \\
	column 5: MAX UNIFORM DEPTH - the maximum allowed depth in the partial depth scheme considering a step of 6 decreased with a value depending to the quality of the 		move  \\
	 column 6: SOLVED OR NOT - 1 if the case has been solved with the parameters from the other columns\\
	 column 7: STEP SIZE - the  number added to the partial depth for each new level of search in case of moves without importance  \\
	The following is the table with the results of the search experiments:
	\begin{center}
		\begin{tabular}{ | l  |  l  |  l  |  l  |  l |  l  |  l  |  l  |   }
			\hline

                            column 1 & column 2 & column 3 & column 4 & column 5 & column 6 & column 7  \\ \hline
		              1       &    20827   &   1              &  17              &     16           &    1           &            6        \\ \hline  
		              2      &     1080           &   1.25              &   8          &   16             &   0            &     6          \\ \hline 
		              3       &     88532           &   1.25              &    12            &   22             &     0          &    6             \\ \hline 
		              4      &      139545          &    1.25             &       12         &   24             &    0           &       6          \\ \hline 
		              5       &      155130          &     1.25            &   14             &   26             &    0           &       6         \\ \hline 
		              6       &   291714       &   1.25                &    14              &  28                &  0             &     6              \\ \hline 
		               7      &      82208          &   1.25              &      16          &      30          &   1            &   6            \\ \hline 
		               8      &   311166             &  1.5               &      12          &     32           &    0           &             6     \\ \hline 
		               9      &     494560           &   1.5              &      13          &     34           &      0         &              6   \\ \hline 
		        	
			\hline
		\end{tabular}
	\end{center}

	\begin{center}
		\begin{tabular}{ | l  |  l  |  l  |  l  |  l |  l  |  l  |  l  |   }
			\hline
		      
                            column 1 & column 2 & column 3 & column 4 & column 5 & column 6 & column 7  \\ \hline

                                     10      &   1009407             &  1.5                &   14                &    36            &    0           &            6      \\ \hline 
		                11     &   208423             &  1.5               &     15           &      38          &     1          &             6    \\ \hline 
		               12      &    1821489            &  1.75               &     13           &   40             &    0           &           6      \\ \hline 
		                 13    &     2337740           &   1.75              &    14            &    42            &    0           &           6       \\ \hline 
			                14   &     381146           &          1.75       &     14           &      44          &     1          &           6      \\ \hline 
		                  15   &     4547933           &  2.00      &     14           &     46           &          0     &           6      \\ \hline 
		                   16  &      603499   &     2.00      &     14       &   48     &      1    &           6        \\ \hline 
		                   17  &     8549650    &        2.25         &    14    &   50    &       0   &           6       \\ \hline 
		                   18  &      816524    &       2.25          &      14     &      52      &      1      &           6      \\ \hline 
		                   19  &         822539  &        2.5           &      14     &    54     &        1       &           6       \\ \hline 
		                    20 &     880194    &      2.75        &      14      &      56       &      1         &            6      \\ \hline 
		                    21 &      897504    &       3    &       14         &   58             &     1          &            6       \\ \hline 
		                    22 &        1026531   &      3.25           &      14          &    60            &     1          &            6      \\ \hline 
		                    23 &        2280040  &  3.5      &    14            &       62         &   1            &            6      \\ \hline 
		                    24 &       96973328    &        3.75         &      14          &    62            &    0           &            6      \\ \hline 
		                    25 &        3210105  &       3.75       &        14     &       64      &      1      &            6      \\ \hline 
		                    26 &         2661590  &      4.00           &     14           &       64         &      1         &            6    \\ \hline 
		                    27 &      4084892     &        4.25         &     14           &      66          &      1         &            6      \\ \hline 
		                    28 &       6624146   &      4.5           &     14           &    68            &       1        &            6     \\ \hline 
		                    29 &           4572359     &   4.75             &   14             &  69              &       1        &            6      \\ \hline 
		                    30 &        7711638    &   5            &    14            &     70          &      1    &   6      \\ \hline

			\hline
		\end{tabular}
	\end{center}

	\subsection{ Interpretation of the experimental results }

	\subsubsection{ (\romannumeral 1)   The step of the search  }

	At first a step representing a fraction of 1 has been used. However, better results have been obtained by using a step bigger than 1 for 
	not so interesting moves. The cause is the decrease in the sensitivity of the output and of other search dependent parameters in regard to the variations of other parameters and of the positional configurations.

	\subsubsection{  (\romannumeral 2)  The importance of detecting decisive moves early  }

	The detection of the variation leading to the objective early decreases the number of nodes searched very much. The fact that the mate has been found at 13 plies depth
	after only 20000 nodes searched shows the line to mate has been one of the first lines tried at each level, even without using knowledge. As it can be seen from the
	table if the mate is detected relatively fast the number of nodes searched is more than 10 times smaller.
	The next plot shows this. The maximums in the number of nodes represents the configurations ( a set of parameters ) for which the mate has not been fast detected.
	The OX represents the number dividing 	the factor giving importance to some significant moves and on OY it is represented the number of nodes searched. \newline

	 Plot of the increase in number of nodes when the importance given to moves with high information gain is decreased \newline         
	
	\includegraphics[width = 2.5in , height = 2.5in]{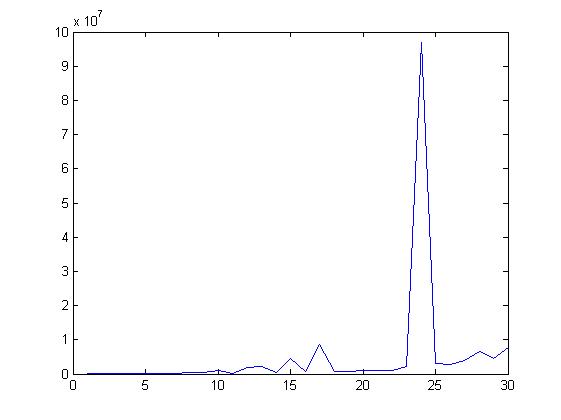} \newline

	On OX it is represented the virtual depth. On OY it is represented the number of nodes.
	
	As it can be seen, even a deeper search that detects the decisive line will explore less nodes than a shallower search that does not find the decisive line.
	For this heuristic and for most of the combinations, when the mate or a strongly dominant line is found fast, the drop in the number of nodes searched is as high 
	as 10 times, even if the uniform search is parametrized for a higher depth.

	\subsubsection{   (\romannumeral 3) The effect of the importance given to high information lines  }
	The number of nodes to be searched increases very much with the decrease of importance given to important moves and to lines of high informational value. 
	
	The following plot, based on data from the previous table shows the increase in the number of nodes explored with the decrease in the importance given to information
	gain when the solution is found. The less importance to the information gaining moves and lines is given, the greater the need for a higher amount of nodes to be searched in order to find the 
	solution.
	
	\includegraphics[width = 2.5in , height = 2.5in]{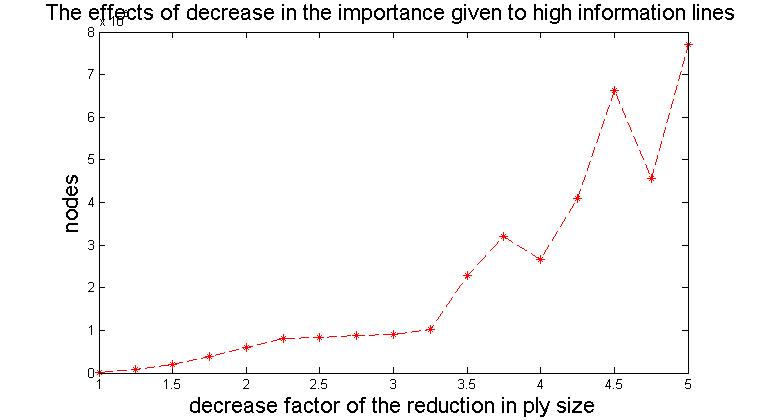}	\newline

	On OX it is represented the TERM DIVIDING THE REDUCTION IN PLY which represents the number dividing the term decreasing the size of the normal ply added to the current 		depth. On OY it is represented the number of nodes.
           The plot shows the explosion of nodes required to find a solution when the importance given to high information lines is decreased. As the importance given to 
	high information lines is decreased the number of nodes searched has to be increased. The importance given to information is decreased so the depth of search must 
	be increased to find the solution. 

	The following plot has the same significance but for the case when the solution is not found. \newline

	\includegraphics[width = 2.5in , height = 2.5in]{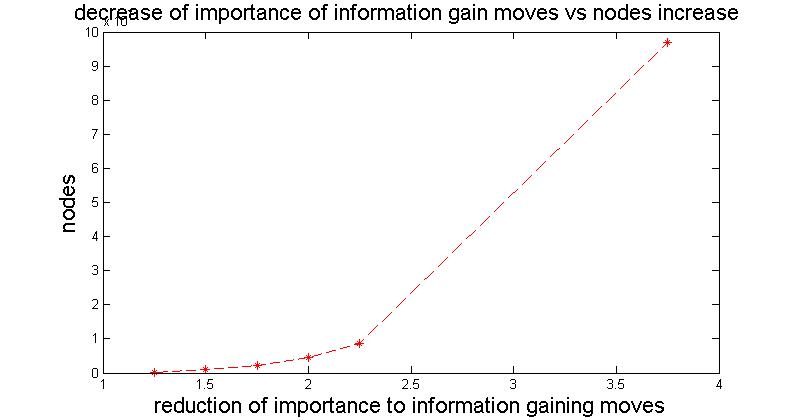} \newline
	
	The plot of nodes searched vs depth when the solution is detected fast shows a far less pronounced combinatorial explosion then when the solution is not found.          The plot shows the explosion of nodes required to find a solution when the importance given to high information lines is decreased. As the importance given to 
high information lines is decreased the number of nodes searched has to be increased.
           It increases even faster when the decisive line is not detected. For a high depth of search, the search cost registers an explosion when no decisive move is found
	reasonably fast.

	When less importance is given to high information gain moves the number of plies has to be increased to compensate this and the number of nodes explodes with the
	number of plies.
	The plot shows the necessary increase of depth when the importance of high information gain moves is decreased.

	For the case when the problem is solved the plot is: \newline

	\includegraphics[width = 2.5in , height = 2.5in]{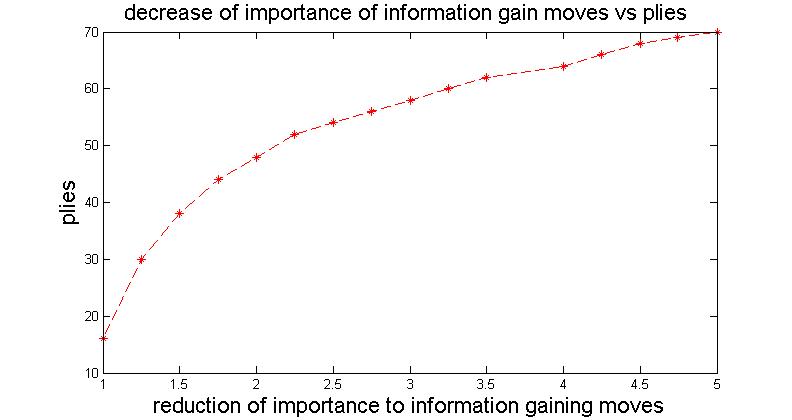} \newline

	Now we can analyze the data for the cases when the solution is not achieved. For the case when the position is not solved is a similar plot but the search at the respective depth has been realized at a far greater cost than when the solution has 			been found fast: \newline

	\includegraphics[width = 2.5in , height = 2.5in]{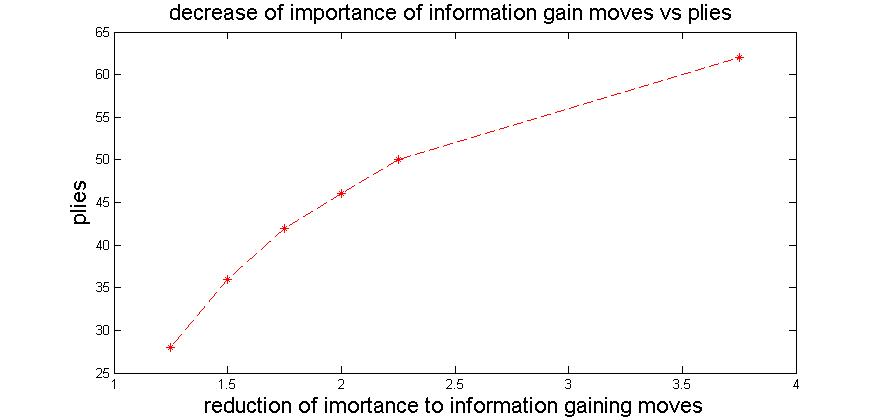} \newline

	\subsubsection{   (\romannumeral 4) The maximum depth and the importance given to high information lines }

	The maximum depth achieved decreases with a decrease in the importance given to areas of the tree with high information.
	Maximum depth vs importance given to information gain. If less importance is given to moves with high information gain more resources are needed for attaining a maximum given 		depth. This is the case for solving some combinations.

	\includegraphics[width = 2.5in , height = 2.5in]{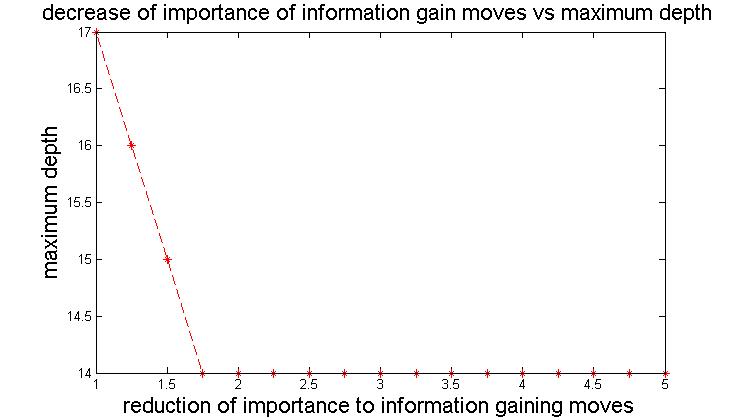} \newline

	As it can be seen from the previous plot, the maximal depth has been achieved also when the solution has not been found but as it can be observed from the above table and plots, at an ever increased cost.

	For the search experiments when the solution has not been found the highest depth remains the same but this time the cost of resources needed to sustain that depth increased very fast, faster than in the previous plot when the solution has been found.

	\includegraphics[width = 2.5in , height = 2.5in]{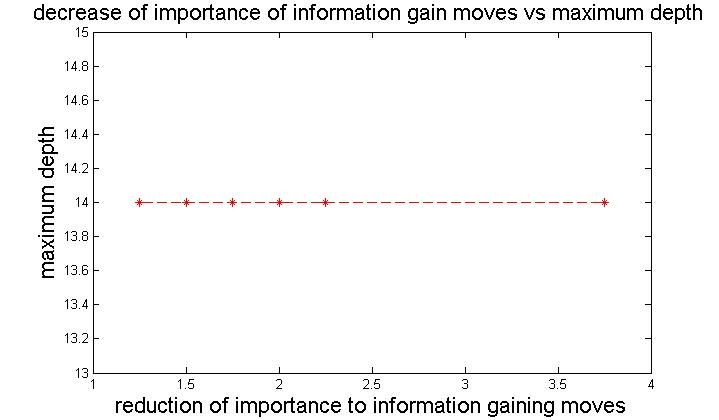} \newline

	\subsubsection{   (\romannumeral 5)  The depth of search and the objective }
	The search detects the mate even if the maximum length is just one ply deeper than the length of the combination.
	Even if we keep the maximum depth constant at far greater cost the searches are less likely to find the decisive lines as it can be seen from plots. 

	\subsubsection{  (\romannumeral 6) The effect of partial depth on the distribution of depths }

	As the search has been changed and less importance has been given to interesting moves, the range in the length of the variations became smaller as less energy
has been allocated for the most informative search lines than previously and more energy to the less informative lines. After shifting ever more resources from 
the informative line to other lines, the objective, the solution of the combination, has not been attained any more by the best lines who did not have the energy this time to penetrate deep enough. The best variations did not have any more the critical energy to penetrate the depth of the state space and solve the problem. The weaker lines
were not feasible as a path for finding any acceptable solution. From this we can understand the fundamental effect of resource allocation. And how marginal shifts in resources
can lead in this context to completely different result. If somebody used the same depth increase for each move, therefore allocating the resources uniformly to the 
variations only a supercomputer can go as deep as it is needed for finding the solution to this combination which is not among the deepest. With the introduction of 
knowledge and heuristics much greater performance would be possible. The experiment concentrates on one heuristic and its effect on the search is highly significant.

	\subsubsection { (\romannumeral 7) On deep combinations  }

	In order to solve deep combinations where some responses are not forced  a program must have chess specialized knowledge 
( or an extension of the information theoretical model of computer chess to all chess theory ) in order to give importance to variations without active moves but with significant tactical maneuvering between forceful moves such as checks and captures.

	\section{Discussion}

	\subsection{General considerations}

	\paragraph{Stochastic modeling in computer chess}

	In the context of game theory, chess is a deterministic game. The practical side of decision in chess and computer chess has many probabilistic elements. The decision is deterministic, but the system that takes the decision is not deterministic, it is a stochastic system. The human decision-making
system and its features such as perception and brain processes are known to be stochastic systems. In the case of computer chess many of the search processes are also stochastic, as it has been seen from the previous examples.

	\subsection{The scope of the results}

The implications of the information theoretic model in terms of heuristic development are discussed in this paper. The extension of the model for more 
elements of computer chess are left for a different research.
	
	\subsection{The limitations of the model}

	The limitations of the model are given by the ability to detect the information gain resulted from different moves and to quantify the information
	gain resulted from these moves.

	\subsection{outlook}

	The objective for future research is to explain also other methods from computer chess using the information theoretic model. Applications also 
	in the case of other strategy games are also a future objective.
	
	\subsection{Conclusion}

		The model starts from the axiomatic framework of information theory and describes in a formal way the role of information in 
	the efficiency and effectiveness of the heuristics used in computer chess and other strategy board games.
	The model proposed considers information in its formal information theoretical meaning as the objective of exploration and the essential factor in the quality of decision in chess and computer chess as well as in other similar games.
 The method of partial depths scheme, well known in practice has
	been described mathematically by observing the fundamental fact that information gain is the criteria that determines the decrease in the uncertainty of the position. The uncertainty of the position is described in a mathematical way through the concept of entropy.
	The information gain  describes in a information theoretic way the decrease in uncertainty resulted from making a move. In this way, a quantification of search information is realized.
	This refers to entropy as it is understood in information theory but it is possible to build parallels also with thermodynamics.
	Previous approaches relied on intuitive formulas and descriptions of the best moves in terms of ''interestingness'' or in terms of chess theory or using knowledge extracted from the games of strong players.
	The approach of the method proposed here is different in that it explains an important method such as the fraction ply method using mathematical methods and formulas that can be derived from the axioms of information theory and determines important coefficients such as the fraction ply associated with moves. The problem of NxN chess is a generalization of the 8x8 chess. It can be expected that the general approach proposed would give a general 
method for the NxN problem where specialized knowledge is not known and would also provide a method to analyze other EXPTIME-COMPLETE problems 
which can be transformed in the NxN chess. The method provides a new understanding of chess, a game analyzed scientifically before by scientists
such as Norbert Wiener, John Von Newumann, Allan Turing, Claude Shannon, Richard Bellman and other famous scientists.
The method proposed generalizes previous approaches and grounds them on information theory a field with a strong theoretical axiomatic system. It can be expected that the method can provide an example on how to quantify search for difficult problems from classes with high complexity and connect 
search in computer science also to physics through the common concept of entropy.

           \paragraph{\emph{Acknowledgment :} The author acknowledges with thanks his discussions with Alberto Giovanni Busetto and 		Prof. J. Buhmann }

	\bibliographystyle{plain}

          \section{APPENDIX}

	\appendix

	\section{Code}

\begin{verbatim}
	double minimax(double alfa,double beta,int depth,int k,int type,move mv,
							double previousval,double virtualdepth){
	move* listNewMoves = (move*) new move [100];
	move mr;    double value = 0 , temp = 0 , ev = 0 ;  int c,number;

	if( (virtualdepth >= maxDepth || depth >= maxExtension ) ){		
		 return evaluation(type,mv); 
	}else{
		if( tip == 1 ){     value = -10000;         }
		else{
			value = 10000;
		}
	
	generator(mv,listNewMoves,number);
	
	for(int i=1; i <= number ;i++){
		listNewMoves[i].eval = fabs( evaluation(tip,listNewMoves[i]) -  previousval ) ;
		double b = -1;
		if( isCheck( listNewMoves[i] ) )
			listNewMoves[i].eval += 10000;	
	}
		
    if( number == 0 ){
		if( tip == 1 )
			if( !is_legal_w(mv) ) return inf_plus;
			else return 0; 
		}else{
			if( !is_legal_n(mv) ) return inf_neg;
			else return 0;
		}
	}else
		for(int k1=1;k1 <= number;k1++){	
			double max = -1;
		int ic = -1;
		for(int c = 1 ; c <= number  ; c++ ){
			double comp = listNewMoves[c].eval;
			if( comp > max ){
				max = listNewMoves[c].eval; 
				ic = c;
			}
		}
	
		mr.eval = listNewMoves[ic].eval;
		double evalPosition = listNewMoves[ic].eval;
		lista_pozitii_urm[ic].eval = -2;
		copy(mr.move, listNewMoves.move);
		copy( mr.tabla,  listNewMove[ic].tabla );
		mr.turn = lista_pozitii_urm[ic].turn;
		double nextV = evaluation(tip,mr);
	
	if(  evalPosition > 2000    )
		value = - minimax( -beta ,-alfa, depth + 1 , ic ,-tip,mr,nextV,virtualdepth );
	else {
			double add = log(fabs(0.1 + (evalPosition/100)))/(log(10.0)) + 5.0/log(number + 2 );		
			value = - minimax( -beta ,-alfa, depth + 1 , ic ,-tip,mr,nextV,virtualdepth + 6 - add);
	}
			if( value >= alfa )   alfa = value;

			if( alfa >= beta ){
				cutoff++;
				break;
			}
	}
	return alfa;
	}
}

\end{verbatim}

\section{Brief description of some chess concepts}

	\paragraph{The reason for presenting some concepts of chess theory. }  
Some of the concepts of chess are useful in understanding the ideas of the paper. Regardless of the level of knowledge and skill in mathematics without a minimal understanding of important concepts in chess it may be difficult to follow the arguments.
  It is not essential in what follows vast knowledge of chess or a very high level of chess calculation skills. However, some understanding of the decision process in human chess, how masters decide for a move is important for understanding the theory of chess and computer chess presented here. The theory presented here describes also the chess knowledge in a new perspective assuming that decision in human chess is also based on information gained during positional analysis. An account of the method used by chess grandmasters when deciding for a move is given in a very well regarded chess book. ~\cite{Kotov}.

\paragraph{Combination } A combination is in chess a tree of variations, containing only or mostly tactical and forceful moves, at least a sacrifice and  		resulting in a material or positional advantage or even in check mate and the adversary cannot prevent its outcome.    The following is the starting position of a combination.

	\includegraphics[width = 2in , height = 2in]{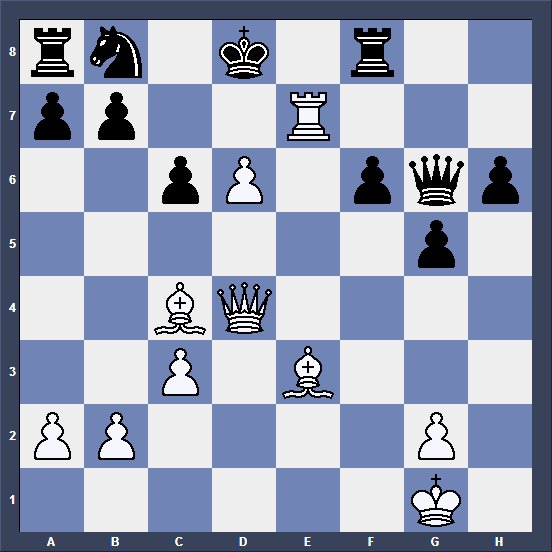}

The problem is to find the solution, the moves leading to the objective of the game, the mate.

	\paragraph{The objective of the game.} 
The objective of the game is to achieve a position where the adversary does not have any legal move and his king is under attack. For example a
mate position resulting from the previous positions is:

	\includegraphics[width = 2in , height = 2in]{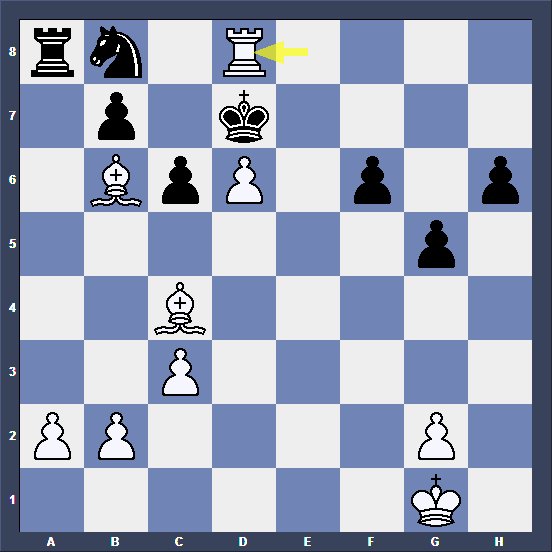}

	\paragraph{The concept of variation } 
A variation in chess is a string of consecutive moves from the current position. The problem is to find the variation from the start position to mate.

In order to make impossible for the adversary to escape the fate, the mate, it is desirable to find a variation that prevents him from doing so, restricting
as much as possible his range of options with the threat of decisive moves. 

	\paragraph{Forceful variation } 
A forceful variation is a variation where each move of one player gives a limited number of legal option or feasible options to the adversary, forcing the adversary to react to an immediate threat.

The solution to the problem, which represents also one of the test cases is the following:

1. Q - N6 ch! ; PxQ 2. BxQNPch ;  K - B1 3. R - QB7ch ; K - Q1 4. R - B7 ch  ;  k - B1  5. RxRch ;  Q - K1     6. RxQch ;  K-Q2      7. R-Q8  mate 

	\paragraph{Attack on a piece } 
In chess, an attack on a piece is a move that threatens to capture the attacked piece at the very next move.  
For example after the first move, a surprising move the most valuable piece of white is under attack by the blacks pawn.

	\paragraph{The concept of sacrifice in chess } A sacrifice in chess represents a capture or a move with a piece, considering that
	the player who performs the chess sacrifice knows that the piece could be captured at the next turn. If the player loses a piece without realizing the piece could be lost then it is a blunder, not a sacrifice. The sacrifice of a piece in chess
	considers the player is aware the piece may be captured but has a plan that assumes after its realization it would place the initiator in advantage or may even win the game. For example the reply of the black in the forceful variation shown is to capture the queen. While this is not the only option
possible, all other options lead to defeat faster for the defending side.  The solution requires 7 double moves or 13 plies of search in depth.

	\section{The axiomatic model of information theory}

	\subsection{Axioms of information theory}

	The entropy as an information theoretic concept may be defined in a precise axiomatic way. ~\cite{CoverThomas}.
	 
	Let a sequence of symmetric functions \( H_m( p_1 , p_2 , p_3, \ldots , p_m ) \) satisfying the following properties:\\
	(\romannumeral 1)  Normalization: \begin{equation}   H_2( \frac{1}{2} , \frac{1}{2} ) = 1    \label{EQ}\end{equation}
	(\romannumeral 2)  Continuity:\begin{equation} H_2(p,1 - p) \label{EQ}\end{equation}   is a continuous function of p \\        
	(\romannumeral 3) \begin{equation} H_m(p_1,p_2,...,p_m ) = H_{m-1}(p_1 + p_2 , p_3,...,p_m) = (p_1+ p_2)H_2(\frac{p_1}{p_1 + P_2},
\frac {p_2}{p_1 + p_2} )	\label{EQ}\end{equation}

	It results $H_m$ must be of the form 

	\begin{equation} H_m = - \sum_{x \in S} {p(x)*\log{ p(x)}}   \label{EQ}\end{equation}

	\section{Contents} 

	\tableofcontents
	
\end{document}